\documentclass[final]{cvpr}
\usepackage{times}
\usepackage{epsfig}
\usepackage{graphicx}
\usepackage{amsmath}
\usepackage{amssymb}
\usepackage{comment}
\usepackage{float}
\usepackage[pagebackref=true,breaklinks=true,colorlinks,bookmarks=false]{hyperref}
\def\cvprPaperID{11394} 
\def\confYear{CVPR 2021}
\begin{document}
\title{Towards Accurate and Robust Classification in Continuously Transitioning Industrial Sprays with Mixup}
\author{Hongjiang Li, Huanyi Shui, Alemayehu Admasu, Praveen Narayanan, \and Devesh Upadhyay \\
Ford Motor Company\\
2101 Village Rd, Dearborn, MI 48124\\
{\tt\small {hli163, hshui, aadmasu, pnaray11, dupadhya}@ford.com}
}
\maketitle
\begin{abstract}
Image classification with deep neural networks has seen a surge of technological breakthroughs with promising applications in areas such as face recognition, medical imaging, autonomous driving.  In engineering problems however, such as high speed imaging of engine fuel injector sprays or body paint sprays, deep neural networks face a fundamental challenge related to the availability of adequate and diverse data. Typically, only thousands or sometimes even hundreds of samples are available for training. In addition, the transition between different spray classes is a “continuum” and requires a high level of domain expertise to label the images accurately. In this work, we used Mixup as an approach to systematically deal with the data scarcity and ambiguous class boundaries found in industrial spray applications. We show that data augmentation can mitigate the over-fitting problem of large neural networks on small data-sets, to a certain level, but cannot fundamentally resolve the issue. We discuss how a convex linear interpolation of different classes naturally aligns with the continuous transition between different classes in our application. Our experiments demonstrate Mixup as a simple yet effective method to train an accurate and robust  deep neural network classifier with only a few hundred samples.
\end{abstract}
\section{Introduction}
Deep convolutional neural networks have led to a series of technological breakthroughs in computer vision applications~\cite{Athanasios2018}. Among the most important factors that contributed to this tremendous success are publicly available, large, high-quality datasets such as ImageNet~\cite{5206848}, CIFAR~\cite{Krizhevsky09learningmultiple}, CelebA~\cite{liu2015faceattributes}. However, we find that pre-trained deep convolutional neural networks face unique challenges when directly applied in scientific domains where datasets are not only very different but also scarce and require expert domain knowledge for accurate labeling and annotation. We also find that data scarcity combined with class overlap naturally leads to overfitting and poor model performance. 

\begin{figure*}
\begin{center}
\includegraphics[width=1.0\linewidth]{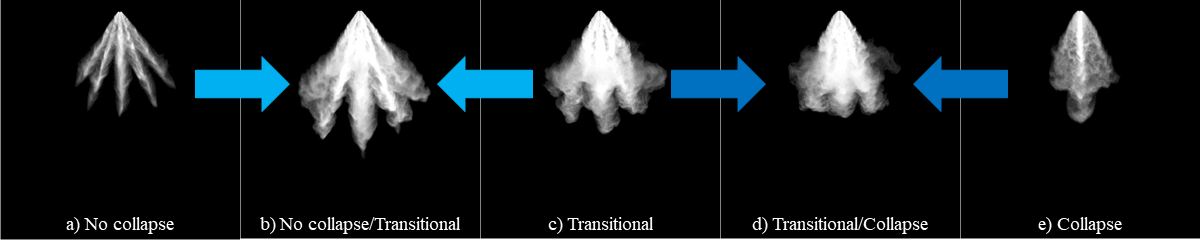}
\end{center}
\caption{Continuous transition between different spray morphology classes: while (a), (c), and (e) show distinct features of the corresponding classes, (b) and (d) show "blending" of two neighbor classes.}
\label{fig:Continuous_Transition}
\end{figure*}

Although many strategies such as less complex models, data augmentation, dropout~\cite{JMLR:v15:srivastava14a}, and regularization can be used to prevent overfitting, their effectiveness in limited datasets containing only hundreds of training samples can be limited. In addition, these methods often lack physical interpretation, as is critical for model acceptance in scientific applications.  
In this work, we applied a deep convolutional neural network to such an engineering problem where high speed images of engine sprays that need to be classified into different categories. A unique challenge in this application is that the transition between different classes is not sharp or perceptibly different, as illustrated in Figure~\ref{fig:Continuous_Transition}, where fuel sprays injected into a constant volume chamber were recorded with Mie-Scattering imaging. Three spray morphology classes can be identified, namely no collapse, transitional, and collapse as shown in (a), (c), and (e), respectively. It is clear that the transition between spray morphology is a "continuum" and some spray images exhibit features from two different classes, as shown in (b) and (d). Those mixed images, represent class overlap and are difficult to label even for domain experts. For example, (b) can be labeled as no collapse, but its close variant from the next camera frame may be labeled as transitional. This unique characteristic unintentionally leads to some "corrupt labels" in the training set. Despite the remarkable capabilities of deep convolutional neural networks, these "corrupt labels" can easily be memorized during the training process, leading to poor real-world model performance.

Another challenge we face is the limited dataset. Due to the limited hardware resources and manpower, we were able to collect and label only 900 images from the spray laboratory. The small dataset further aggravates the incidence of "corrupt labels" due to class overlap thereby impacting model performance. To systematically deal with these two challenges in a physically meaningful manner, we chose the Mixup approach for data augmentation as proposed by Zhang \etal~\cite{DBLP:journals/corr/abs-1710-09412}. Our study suggests that the convex linear interpolation of Mixup naturally aligns with the continuous class transition observed in our dataset. Depending on whether the two candidate images belong to same class or not, Mixup can either function as a data augmentation technique that reduces overfitting or a label smoothing technique that mimics the continuous class transitions. Our study also shows that even though data augmentation can mitigate overfitting to some degree, it does not bring additional benefits and may negate the performance boost if stacked with Mixup.  

The key contributions of our work are listed below:

\begin{itemize}
    \item We propose an explainable training technique for robust and accurate deep convolutional neural network classifier suitable for classifying continuously transitioning industrial sprays.
    \item We showcase that Mixup works well with limited datasets containing only 900 samples, this benefit is not discussed in the original work by Zhang \etal~\cite{DBLP:journals/corr/abs-1710-09412}.
    \item We demonstrate that Mixup can expand training distributions to mimic the "continuum" of class overlap observed in our dataset, hence resolving the overfitting issue and leading to good performance in real-world testing.
\end{itemize}
\section{Background and Motivation}
Although the automotive industry appears to be on the verge of transitioning to electric vehicles, the vast majority of passenger cars and commercial vehicles on the road are still powered by Internal Combustion Engines (ICE). ICEs are complex systems that convert the energy stored in the hydro-carbon bonds of chemical compounds in fossil fuels to mechanical energy used to power vehicles. A basic ICE works like follows: the fuel system injects gasoline (or diesel) into the intake port (or combustion chamber) via high-pressure fuel injectors, the fuel then evaporates and mixes with the induced fresh air, the fuel-air mixture ignites by spark plug(or auto-ignites by compression), creating a high temperature, high pressure explosion that provides motive power.  

The fuel injection and mixing with ambient air is one of the most important factors impacting engine performance (power) as well as engine emissions (such as CO2, and NO). Therefore, extensive studies have been carried out in the combustion domain to optimize fuel sprays for maximizing the power to emissions index.  Among many measurement techniques used in spray testing, Mie scattering remains one of the prominent imaging methods for spray visualization~\cite{todd2014}. Mie scattering uses a light source (e.g., laser) and a camera to record the macroscopic spray development inside a quiescent chamber filled with pressurized air. Light is elastically scattered by fuel droplets similar to or larger than the wavelength of the incident light. The signal collected by a camera is proportional to the integral of cubic of droplet diameters along the line of sight. Examples of Mie scattering imaging are shown in Figure~\ref{fig:Continuous_Transition}, where the dark color is background and the light color are liquid sprays. The grayscale in each image roughly indicates the intensity of liquid volume fractions.

One of many insights gained from Mie scattering is spray morphology which helps engine experts to understand the mechanisms of spray breakup and mixture formation. However, spray development is a complicated process affected by many factors such as fuel volatility, fuel temperature, injector geometry, ambient conditions, and turbulence. Despite the continuous efforts in academia and industry, spray development is still not fully understood. It is generally accepted that, depending on the macroscopic features of spray images, spray morphology can be classified into three regimes, namely no collapse, transitional, and collapse. Characteristics of each regime are listed blow:  

\begin{itemize}
    \item No collapse regime: Characterized by visually discernible narrow plumes or branches. The separations of spray plumes occur very close to the injector tip located on the top. 
    \item Transitional regime: Characterized by wide spray plumes that begin to interact with each other. The separations of plumes move downstream and the exact locations becomes hard to visually discern. The spray structure still resembles a cone shape.
    \item Spray collapse regime: Characterized by one single prolonged central plume. There may be one or two discernible spray plumes but the majority of them are coalesced.
\end{itemize}

It is widely accepted that spray collapse should be avoided at all cost since it can lead to spray impingement, reduced total surface area for fuel-air mixing, and poor atomization. The combined effects of these are worsened engine performance and increased combustion emissions that may prevent the engine from mass production due to stringent emission standards. Unfortunately, the spray collapse phenomenon is not well understood and prediction of spray collapse is very challenging and has so far remained a manual process requiring many hours of subjective evaluations by domain experts. This challenge is our motivation to use deep convolution neural networks for automating the robust classification of spray morphology and detection of spray collapse.
\section{Related Work}
\subsection{Physics Domain}
Despite its detrimental effects on ICE performances, spray collapse or spray morphology classification remains a difficult problem.  This challenge is further exacerbated from design changes such as the move from outward opening swirl injectors to multi-hole direct injection injectors. With recent advances in turbo-charging, downsizing, and new injection strategies, domain experts are increasingly relying on experiments or high-fidelity simulations to predict spray structures, especially the transition to spray collapse. Many of the reported work focus on developing experimental correlations based on carefully designed experiments. However, due to the limitation of hardware and resources, the reported correlations rarely give satisfactory results on conditions out of the original research scope. Some of the works even reported contradictory findings. Zeng \etal~\cite{ZENG2012287} found that the spray morphology transition can be solely predicted by the ratio of ambient pressure to fuel saturation vapor pressure. The tested spray stays in the transitional regime when the pressure ratio is between 0.3 to 1.0, and spray collapse happened when its value is less than 0.3. On the contrary, Lacey \etal~\cite{LACEY2017345} reported that the pressure ratio was not effective at predicting the transition to spray collapse, especially of different fuels. Various theories~\cite{Mojtabi2008THEEO, ALEIFERIS2013143, WU2016322} have been proposed in the literature, but a universally applicable model has yet to be discovered. 

It should be noted that we do not criticize the findings in ~\cite{ZENG2012287, LACEY2017345, Mojtabi2008THEEO, ALEIFERIS2013143, WU2016322}. In fact, we believe each model or theory is perfectly explainable to the researchers within the scope of their studies. However, each study represents an enormous effort in the design and conduct of the experimental measurements, which inevitably, resulted in a limited dataset for analysis. Therefore, those findings are limited in the sense that they can only provide accurate information within the data space explored by each corresponding study.   
\subsection{Machine Learning Domain}
Though deep neural networks have found various applications in the scientific domain such as medical imaging, materials discovery, computational fluid dynamics to name a few. To the authors' best knowledge there are no efforts in the classification of continuously transitioning spray structures. On the training front new methods and/or training techniques have been reported since the original work of Mixup by Zhang \etal~\cite{DBLP:journals/corr/abs-1710-09412}, which demonstrated that Mixup reduces test errors of multiple state-of-the-art deep convolutional neural network models on ImageNet, CIFAR, and speech data. Berthelot \etal~\cite{DBLP:journals/corr/abs-1905-02249} expanded the idea to semi-supervised learning (SSL) and proposed a "holistic" learning method named MixMatch. Their experiments on SSL suggest that MixMatch significantly improved performance compared to other methods they studied. The same group took a step further and proposed the ReMixMatch by introducing augmentation anchoring and distribution alignment to MixMatch~\cite{DBLP:journals/corr/abs-1911-09785}. Their experiments show that this SSL algorithm can reach or beat MixMatch with much less data. Jeong \etal~\cite{DBLP:journals/corr/abs-2006-02158} used Mixup as one of the functioning elements and proposed an Interpolation-based Semi-supervised learning method for object detection (ISD). They demonstrated that ISD significantly improves the performance of Single Shot Multibox Detector(SSD)~\cite{2016SSD} in both supervised and semi-supervised object detection tasks.

\section{Data and Method}
\subsection{Training data}
Due to the nature of our problem, the entire dataset is composed of proprietary spray images collected from our internal spray laboratory. We wish to develop a deep learning model that is capable of classifying spray structures over the various testing conditions, injection timings, and experimental setups. This allowed spray images to be collected from multiple sources, covering a wide range of operating conditions, injector geometries, and measurement events. Another consideration during the data collection process is that we wish to cover the entire injection event, so an additional class named "Pre/Post" (short for pre-injection and post-injection) was added to the dataset so the final model can take any input from the Mie Scattering without human attention.

\begin{figure*}
\begin{center}
    \includegraphics[width=1.0\linewidth]{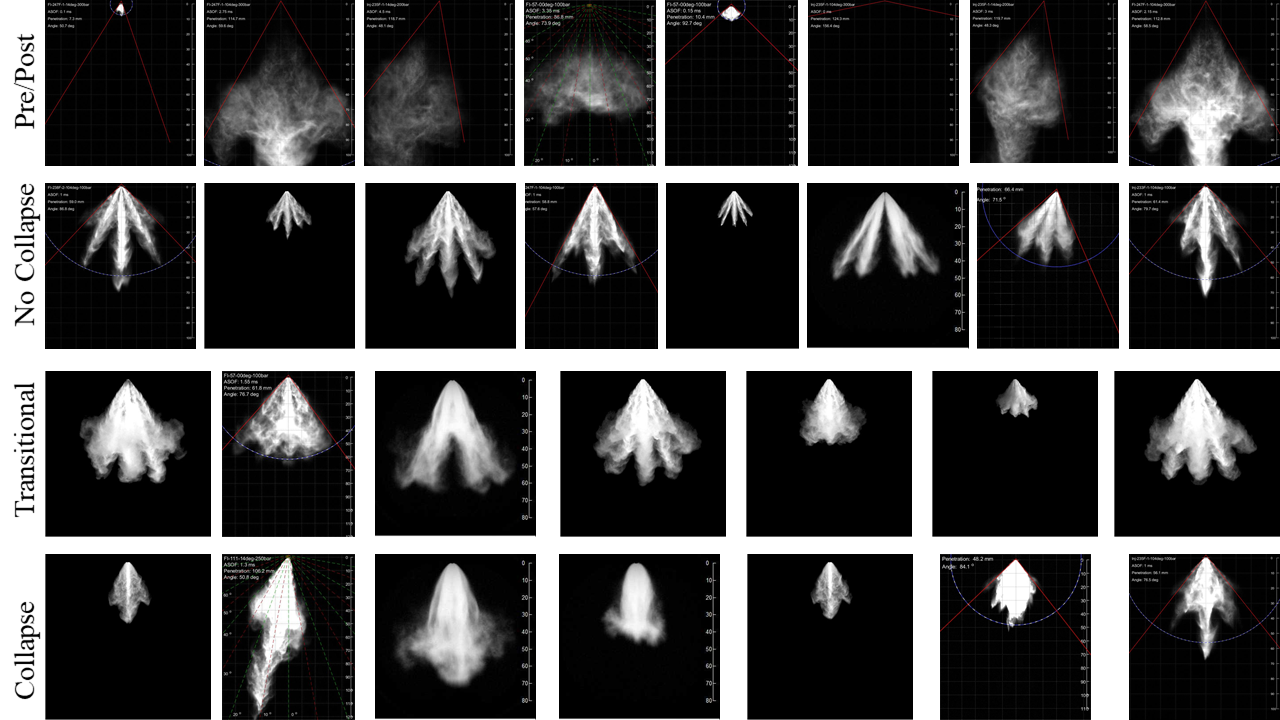}
\end{center}
\caption{Examples of dataset used for training convolutional neural networks models. An additional class, namely "Pre/Post" was added to the dataset to cover the entire injection event. Note that some examples shown in this figure have text annotations, solid red or dotted white boundary lines from the processing tool of Mie Scattering. During the data collection and labeling process, we did not remove these "noises" for the sake of training a robust model.}
\label{fig:HowTheDataLookLike}
\end{figure*}

Table~\ref{tab:dataDistribution} lists the number of examples for each class, giving a total of 878 images. We set aside 75\% of the dataset for training, 15\% for validation, and 10\% for testing, leaving a total of 658, 152, and 88 images for training, validation, and testing, respectively. Examples of spray images in each class are shown in Figure~\ref{fig:HowTheDataLookLike}. Since the data were collected from multiple sources, they are not necessarily the same size as can be seen in Figure~\ref{fig:HowTheDataLookLike}. In addition, some of the images may carry auxiliary boundary lines and/or text annotations processed by Mie Scattering tool. Those lines were not removed during the labeling process and we anticipate the model would either treat them as noise or recognize them as useful features.

\begin{table}
\begin{center}
\begin{tabular}{|l|c|}
\hline
    Class           & \# of examples\\
\hline\hline
    Pre/Post        & 173           \\
    No collapse     & 199           \\
    Transitional    & 241           \\
    Collapse        & 265           \\
\hline
\end{tabular}
\end{center}
\caption{Distribution of spray images among all classes. Note that the dataset is quasi-balanced.}
\label{tab:dataDistribution}
\end{table}

\subsection{Network Architecture}\label{sec:modelSelection}
\begin{figure}[t]
\begin{center}
    \includegraphics[width=0.8\linewidth]{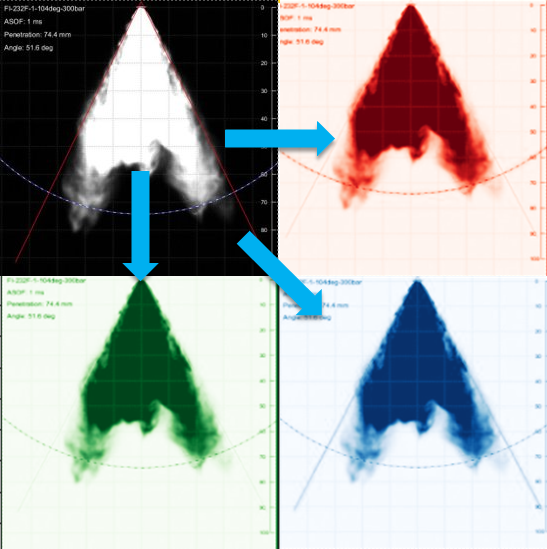}
\end{center}
\caption{Loading grayscale spray image in RGB mode leads to three nearly identical channels. This makes transfer learning with many state-of-the-art models feasible in our application. Note that the colored boundary processed by the Mie Scattering shows strong trace in R and B channels but nearly invisible in G channel.}
\label{fig:RGB_Square}
\end{figure}

Given the limited dataset, the most feasible starting point is transfer learning by reusing the lower layers of pre-trained models. Fortunately, many state-of-the-art deep learning models are released for Keras~\cite{chollet2015keras}. TensorFlow version 2.4.1~\cite{tensorflow2015-whitepaper} is used throughout the work. Among the many available pre-trained models on Keras, we chose ResNet~\cite{DBLP:journals/corr/HeZRS15} family as the starting point after an internal evaluation. All the images were pre-processed as $224 \times 224$ pixel images as expected by ResNet. In addition, the grayscale images were loaded in Red, Green, and Blue channels. As shown in Figure~\ref{fig:RGB_Square}, loading grayscale images in RGB channels leads to three nearly identical colored images. These were then fed into pre-trained ResNet models and fine tuned. 

To select a suitable model for fine tuning, we performed 5-fold cross validation using all 878 available spray images. Seven variants of ResNet models were tested. Note that we implemented ResNet34 from scratch since it is not available from Keras. For the remaining six models, we reused all the pre-trained lower and mid layers and only made the last block (namely, \textit{conv5 block3}) trainable. This leaves about 4.4 million trainable parameters for each model except for ResNet34. We also removed the fully connected top layer and replaced it with global average pooling layer, followed by a dense output layer with one neuron per class. For pre-trained ResNet models, we used Adam optimizer~\cite{Adam} and fixed the learning rate, batch size, and total epochs to be 0.001, 32, and 100, respectively. For ResNet34, we used stochastic gradient descent with a learning rate of 0.0001 and momentum of 0.9.

The ensemble-averaged test accuracies and standard deviations are reported on Table~\ref{tab:5FoldTestAccuracy}. ResNet50 outperforms all other models with an impressive test accuracy of 96\%, though other models are not far behind. Although ResNet101 and larger models have more representation power than ResNet50, they are prone to overfitting with small training datasets and therefore have slightly poor performance on the test set. On the other hand, ResNet34 seems too shallow to learn all necessary low-level and/or high-level features. Given its highest accuracy, ResNet50 is used as the base model for further experiments. 
\begin{table}
\caption{Ensemble-averaged test accuracies and standard deviations of seven ResNet models for the 5-fold cross-validation. Note that ResNet34 was implemented from scratch while the other six were pre-trained models from Keras.}
\begin{center}
\begin{tabular}{|l|c|c|}
\hline
    Model           & Average test accuracy &   STD                 \\
\hline\hline
    ResNet34        & 0.9522                &   0.0142              \\   
    ResNet50        & \textbf{0.9602}       &   0.0196              \\
    ResNet50V2      & 0.9488                &   0.019               \\
    ResNet101       & 0.9556                &   0.0213              \\
    ResNet101V2     & 0.9351                &   0.0284              \\
    ResNet152       & 0.9476                &   0.018               \\
    ResNet152V2     & 0.9385                &   0.0149              \\
\hline
\end{tabular}
\end{center}
\label{tab:5FoldTestAccuracy}
\end{table}
\subsection{Mixup and Data Augmentation}
\begin{figure}[!htbp]
\begin{center}
    \includegraphics[width=0.9\linewidth]{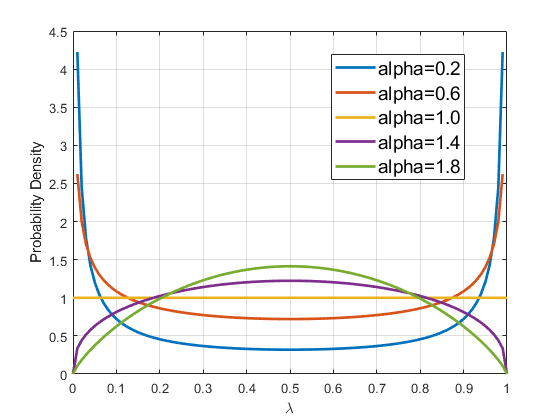}
\end{center}
\caption{The interpolation coefficient $\lambda$ is beta-distributed $\lambda \sim \mathbf{Beta}(\alpha, \alpha)$. Note that as $\alpha$ increases, a realization of $\lambda$ would have a higher chance of being close to 0.5.}
\label{fig:BetaDistribution}
\end{figure}

Mixup is a simple and data-agnostic method proposed by Zhang \etal~\cite{DBLP:journals/corr/abs-1710-09412}. Suppose we have two input image vectors, $x_1$ and $x_2$, and their corresponding one-hot vectors are $y_1$ and $y_2$, then the Mixup augmented training image vectors are given by:

\begin{equation} \label{eq1}
\begin{split}
\Tilde{x} & = \lambda x_1 + (1.0 - \lambda) x_2 \\
\Tilde{y} & = \lambda y_1 + (1.0 - \lambda) y_2
\end{split}
\end{equation}
where $\lambda$ is the interpolation coefficient randomly drawn from the Beta distribution, $\lambda \sim \mathbf{Beta}(\alpha, \alpha)$. As shown in Figure~\ref{fig:BetaDistribution}, $\lambda$ approaches 0 or 1.0 as $\alpha \rightarrow 0$, this essentially eliminates interpolation and selects one input image as the output. As $\alpha$ increases, a realization of $\lambda$ would have a higher chance of being close to 0.5, leading to strong blending of both input images.

There are many choices of Mixup implementations as suggested by Zhang \etal~\cite{DBLP:journals/corr/abs-1710-09412}. For example, the augmented one-hot vector can be the same as the one input image with larger weight. Mixup can also be performed on more than two images. In this work, we implemented the vanilla version given by Equation~\ref{eq1}. 

\begin{figure}
\begin{center}
    \includegraphics[width=1.0\linewidth]{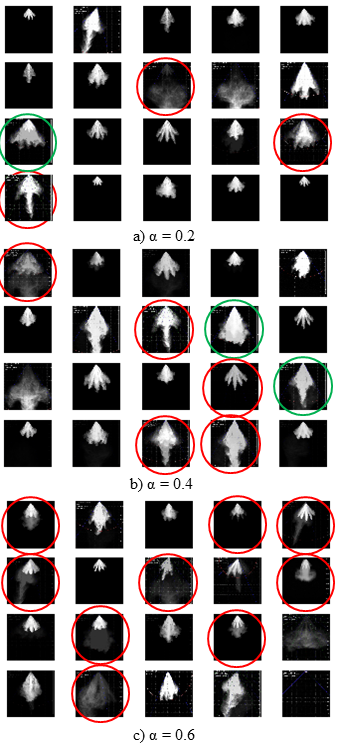}
\end{center}
\caption{Examples of training images after Mixup data generator. Note that both the frequency and strength of interpolation increase as $\alpha$ increases from 0.2 to 0.6.}
\label{fig:MixupDataGenerator}
\end{figure}

Figure~\ref{fig:MixupDataGenerator} shows some examples after Mixup with different $\alpha$ values. Note that we only tested $\alpha$ up to 0.6 as Zhang \etal~\cite{DBLP:journals/corr/abs-1710-09412} found Mixup only improved performance over traditional data augmentation with $\alpha < 0.4$. For larger $\alpha$ values, Mixup leads to underfitting. At $\alpha = 0.2$, four out of 20 randomly selected images are visually discernible as being augmented by Mixup. Among those four images, one of them was interpolated between two images belongs to the same class as highlighted by green cycle. The resulting image expands the data distribution of that class and hence Mixup works like a data augmentation tool. The other three were interpolated between two different classes, the resulting labels are no longer one-hot vectors. This is considered crucial to our application as those cross-class augmented images mimic the smooth transition observed in our dataset. As $\alpha$ increases, the  interpolation becomes stronger, and more images can be observed as being interpolated. 

Data augmentation is widely used to combat overfitting of deep neural networks. However, among many choices of data augmentation methods such as shifting, rotation, flip, zooming, one must be careful as the effectiveness of data augmentation is dataset dependent and use of domain knowledge is usually needed. Although Mixup was found to improve performance over data augmentation~\cite{DBLP:journals/corr/abs-1710-09412}, they can easily work together. In this work, Mixup is performed before any optional data augmentation since it guarantees that there will be only one injector tip located on the top of each augmented spray image. If image augmentation such as rotation and shifting are applied before Mixup, then the Mixup augmented image may have two injector tips, which from the physics point of view pollutes the dataset and will have negative effects on the model performance.   

\section{Experiments}
\subsection{Data Augmentation}
At this stage, we introduce some data augmentations without Mixup to fine-tune the ResNet50 model. Like section~\ref{sec:modelSelection}, only the last block of ResNet50 (i.e., \textit{conv5 block3}) is allowed to be retrained on our dataset. The lower- and mid-level layers from Keras ResNet50 model pre-trained on ImageNet are reused. Random rotation up to 20 degrees is used since in our experiments, the injector may be installed at a slightly angled position. Random shifting is limited up to 20\% as the spray image is almost always centered at the camera window. Horizontal flip is allowed as it produces an image taken at the opposite direction. Vertical flip is not used since it gives an upside-down image that makes no physical sense. Brightness, saturation, hue, or contrast adjustments are excluded as well given Mie Scattering imaging produces consistent images without much distortion. We use Adam optimizer~\cite{Adam} with an initial learning rate of 0.001, and it is then divided by 10 after no improvements observed in validation loss for 5 consecutive epochs. We also use early stopping with a patience of 50 to avoid over-training. In all our experiments, the model will stop training with less than 200 epochs.

The ensemble-averaged training-validation accuracy gaps over the last 50 epochs are reported on Table~\ref{tab:EXPDataAug}, also reported are the test accuracies of the final model. Compared to the baseline without any data augmentation, all single data augmentations lead to improved generalization. Combining two or three augmentation methods do not bring the training-validation accuracies any closer. Another observation is that the test accuracy saturates with all experiments. Given we only have 88 test images, 94\%, 96.5\%, and 97.7\% test accuracies indicate there are only 5, 3, and 2 images being miss-classified by the model. Note that those miss-classified images are not necessarily the same between each test. Although there is no universal guideline as how small the training-validation gap should be to indicate a good model, we argue the value should be at least less than 1.0\% for this application.

\begin{table}
\caption{Training-validation gaps averaged over the last 50 epochs for ResNet50 with different data augmentation methods.}
\begin{center}
\begin{tabular}{|l|c|c|}
\hline
    Model                   &  Gap                      &   Test Acc.           \\
\hline\hline
    Baseline                & 4.9\%                     &   94\%                \\   
    Rot 10                  & 3.9\%                     &   96.5\%              \\
    Rot 20                  & 4.3\%                     &   96.5\%              \\
    Shift 0.1               & 1.6\%                     &   96.5\%              \\
    Shift 0.2               & \textbf{1.3\%}            &   96.5\%              \\
    H. Flip                 & 3.5\%                     &   97.7\%              \\
    Shift 0.2/H. Flip       & \textbf{1.3\%}            &   96.5\%              \\
    Rot 10/H. Flip          & 5.3\%                     &   96.5\%              \\
    Shift 0.2/Rot10         & 2.2\%                     &   96.5\%              \\
    Shift 0.1/Rot10/H. Flip & 2.8\%                     &   96.5\%              \\
\hline
\end{tabular}
\end{center}
\label{tab:EXPDataAug}
\end{table}
\subsection{Mixup}
We evaluate Mixup without data augmentation and summarize the results on Table~\ref{tab:MixupEXP}. Note that all Mixup tests lead to improved performance over data augmentation. The training-validation accuracy gaps are below 1.0\% with all three $\alpha$ values as we desired. Larger $\alpha$ leads to better generalization, but the model under-fits the training set as evidenced by the gradually decreasing ensemble-averaged training accuracies. This finding is consistent with the work by Zhang \etal~\cite{DBLP:journals/corr/abs-1710-09412}, although they use much bigger datasets (ImageNet).

Zhang \etal~\cite{DBLP:journals/corr/abs-1710-09412} introduced "corrupted labels" to CIFAR dataset by replacing up to 80\% of total image labels with random noises. Their testing showed that Mixup can mitigate the memorization of those corrupted labels. In our application, some "corrupted labels" are unintentionally introduced because of the blurry boundaries between continuous class transitions. We hypothesize that Mixup can help combat the memorization of those "corrupted labels" as well. For example, if one image within the "collapse/transitional" is labeled as "collapse" and its neighbor from the next camera frame is labeled as "transitional", then a neural network without Mixup would learn to memorize those labels and over-fit the data. On the other hand, all training images are generated from Mixup (although not all of them are interpolated), so those two images may produce a new training image with an interpolated label. This is equivalent to labeling the newly generated image as "collapse/transitional", thereby reduces the memorization of "corrupted labels". From the physics point of view, this interpolation also aligns with the actual transition in the experiments.       

We also perform a test with Mixup ($\alpha=0.2$) followed by the "best-practice" data augmentation reported on Table~\ref{tab:EXPDataAug}, i.e., shifting by 20\% plus horizontal flip. The training accuracy drops to 93.7\% and the test accuracy is lower than all Mixup-only tests summarized on Table~\ref{tab:MixupEXP}. This indicates additional data augmentation may add excessive regularization. Among all the Mixup tests, we find $\alpha=0.2$ leads to the best model.

\begin{table}[!htbp]
\caption{Training-validation accuracy gaps averaged over the last 50 epochs for ResNet50 with Mixup.}
\begin{center}
\begin{tabular}{|l|c|c|c|}
\hline
    Model               & Training Acc.             &  Gap                      &   Test Acc.           \\
\hline\hline
    Baseline            & 99.7\%                    & 4.9\%                     &   94\%                \\   
    $\alpha = 0.2$      & 96.5\%                    & 0.9\%                     &   98\%                \\
    $\alpha = 0.4$      & 95.1\%                    & 0.3\%                     &   98\%                \\
    $\alpha = 0.6$      & 94.3\%                    & 0.3\%                     &   98\%                \\
\hline
\end{tabular}
\end{center}
\label{tab:MixupEXP}
\end{table}
\subsection{Application}
We perform an additional test of the final Mixup model with $\alpha=0.2$ on a real-world dataset containing 7,200 spray images. Unlike the training set where spray images are collected from multiple sources, this dataset only contains images from one test, albeit multiple testing conditions were covered. The classification results are manually examined by engine spray experts and performance measures are reported on Table~\ref{tab:FinalTesting}. Note that this dataset does not contain any image in the Pre/Post class so the corresponding measures are not available. 

About 64\% of the tested spray images belong to the no collapse regime and they are all correctly detected by the classifier. This is crucial to the engine experts as no collapse sprays are desired for optimal engine performance and emissions, and hence correctly detecting this regime would allow them to use the corresponding testing conditions (ambient pressure, fuel temperature, injector geometry, etc.) for further optimization. On the other hand, only 12.7\% of the test images belong to the collapse regime and the model is able to accurately predict 96\% of them. The precision score of collapse is relatively low (97\%) as well, indicating the model struggles with some images that possibly belong to the "transitional/collapse" regime shown in Figure~\ref{fig:Continuous_Transition} (d). Nonetheless, the model reaches an overall accuracy of 98.7\%, which is 0.7\% higher than the test accuracy on Table~\ref{tab:MixupEXP}.  

\begin{table}[!htbp]
\caption{Precision, recall, and F1 score after applying the final ResNet50 to 7,200 spray images.}
\begin{center}
\begin{tabular}{|l|c|c|c|c|}
\hline
    Class                   &  Precision        &   Recall       & f1-score &Support    \\
\hline\hline
    Pre/Post                & N/A               &   N/A          & N/A      & 0         \\   
    Collapse                & 0.97              &   0.96         & 0.96     & 915       \\
    Transitional            & 0.99              &   0.99         & 0.99     & 1754      \\
    No collapse             & 0.99              &   1.0          & 0.99     & 4531      \\
\hline
\end{tabular}
\end{center}
\label{tab:FinalTesting}
\end{table}
\section{Conclusion}
Data scarcity is one of the main challenges that large deep neural networks face when applied to problems in the scientific domain. In this paper, we showcased a robust training method for an industrial engine injector spray classification problem where the transition between classes is a "continuum". By using Mixup, we were able to train a ResNet50 model with only a few hundred images. The final model achieved 98.7\% prediction accuracy on a real-world spray dataset. Through testing, we found that Mixup improved performance over data augmentation methods. Mixup also provides the benefit of reducing the memorization of "corrupted labels" unintentionally introduced during the labeling process. As we understand, the linear interpolation of both data and labels also agrees with the "continuum" nature of class transitions in injector sprays. 

As future work, we are interested in incorporating simulation data from high-fidelity computational fluid dynamics (CFD) tools into the experimental dataset and continue to explore the properties of Mixup and variants. Simulation data, generally, has more details compared to  Mie Scattering imaging. We also wish to understand the impact of Mixup on assisting these models to grasp knowledge embedded differently in simulation and experiment datasets. 
{\small
\bibliographystyle{unsrt}
\bibliography{egbib}
}
\end{document}